\documentclass[sn-mathphys-num]{sn-jnl}

\usepackage[utf8]{inputenc}
\usepackage{graphicx}
\usepackage{physics}
\usepackage{hyperref}
\usepackage{xcolor}
\usepackage{lineno}

\begin{document}


\title{Autoregressive model path dependence near Ising criticality}

\author*[1]{Yi Hong Teoh}\email{yhteoh@uwaterloo.ca}
\author[1,2]{Roger G. Melko}\email{rmelko@perimeterinstitute.ca}

\affil[1]{Department of Physics and Astronomy, University of Waterloo, 200 University Ave. West, Waterloo, Ontario N2L 3G1, Canada}
\affil[2]{Perimeter Institute for Theoretical Physics, Waterloo, Ontario N2L 2Y5, Canada}


\abstract{
Autoregressive models are a class of generative model that probabilistically predict the next output of a sequence based on previous inputs. 
The autoregressive sequence is by definition one-dimensional (1D), which is natural for language tasks and hence an important component of modern architectures like recurrent neural networks (RNNs) and transformers.
However, when language models are used to predict outputs on physical systems that are not intrinsically 1D, the question arises of which choice of autoregressive sequence -- if any -- is optimal.
In this paper, we study the reconstruction of critical correlations in the two-dimensional (2D) Ising model, using RNNs and transformers trained on binary spin data obtained near the thermal phase transition.
We compare the training performance for a number of different 1D autoregressive sequences imposed on finite-size 2D lattices.
We find that paths with long 1D segments are more efficient at training the autoregressive models compared to space-filling curves that better preserve the 2D locality.
Our results illustrate the potential importance in choosing the optimal autoregressive sequence ordering when training modern language models for tasks in physics.
}

\maketitle


\section{Introduction}\label{intro}
The field of artificial intelligence (AI) has seen a step change in the abilities of a number of unsupervised machine learning architectures over the last year or two.  
In particular, generative pre-trained transformers (GPT) and related language models have come of age, displaying unprecedented accuracy for problems such as machine translation, speech recognition, text generation, and more \cite{openai2024gpt4, devlin2019bert, raffel2020exploring, touvron2023llama}.
The performance of these models on natural language tasks leads to the question of whether other fields of study, particularly those in the physical sciences, could benefit from adopting modern AI strategies into their research toolboxes \cite{LMquantum}. Strong-correlation physics, e.g.~statistical mechanics, condensed matter and quantum information, has benefited widely from this shifting paradigm, with language models aiding in a number of recent transformative results \cite{viteritti2023transformer, zhang2023transformer,Wang2023transformer, bausch2023learning}.
Conversely, the field of strong-correlation physics also provides a  toolset for helping to understand the theoretical underpinnings of AI architechtures such as language models.  For example, since 2016 \cite{Torlai_thermo,CarleoTroyer,QRBM} a number of works connecting restricted Boltzmann machines (RBMs) to physics have used well-established notions surrounding locality, correlations, symmetries, etc.~to advance the theoretical understanding of RBMs \cite{RBMreview}. 
Similar research connecting machine learning to strongly-correlated many-body physics could hold the key to  understanding the success of other architectures, and could provide guidance to improve the performance of language models and other AI strategies.

Consider generative models where the input is mapped to a sequence. Such sequence mapping is obvious for many language tasks such as text completion, where words in a sentence have a naturally occurring order. 
Thus a number of generative models have been developed with the purpose of inferring a probability distribution from a set of unlabeled sequence data. Models where the resulting likelihood is decomposed into conditional probabilities for each sequence element are called {\it autoregressive}.
Examples include NADEs \cite{benigno2016nade}, RNNs \cite{sherstinsky2020fundamentals}, and transformers (which are autoregressive models when using masked attention) \cite{vaswani2017attention}.
Such models are increasingly being used for physical data that is not naturally associated with a sequence, such as for variables residing higher dimensional ($D>1$) lattices or more arbitrary graphs.
In particular, autoregressive models are capable of encoding complex quantum many-body states, where long-range interactions, or non-local correlations such as entanglement, negate any strict 1D sequential interpretation \cite{mohamed2020recurrent,Iouchtchenko2023neural,moss2023enhancing, Cha2022attention, czischek2022data,sprague2023variational, Di2022Autoregressive}.
Thus, there is a need to understand how the decomposition of higher-dimensional lattices or graphs into sequences affects a model's performance. 

In this paper, we study the learning behavior of two leading autoregressive architectures, RNNs \cite{mohamed2020recurrent,Iouchtchenko2023neural,moss2023enhancing,czischek2022data} and transformers \cite{Cha2022attention,sprague2023variational,RydbergGPT,AttentionKim}, when trained on data that is not intrinsically a 1D sequence.  We focus on data obtained from the 2D Ising ferromagnet on a square lattice, a paradigmatic model in statistical physics, with Hamiltonian,
\begin{equation}
\label{Ising}
H = -  \sum_{\langle i j \rangle } \sigma_i \sigma_j,
\end{equation}
where each spin variable $\sigma_i$ is $\pm 1$, and ${\langle i j \rangle }$ denotes nearest-neighbor bonds.
Near its well-studied thermal critical point, 
Ising variables develop long-range correlations that decay as a power law.
By mapping lattice sites to sequence elements, we train several autoregressive models on this data, and observe their behavior in regard to learning optimal model parameterizations.  We find that training dynamics depends strongly on the arbitrary choice of autoregressive path through the 2D lattice, and discuss possible consequences for improving training behavior of large autoregressive models in the future.


\section{Autoregressive generative models}\label{model}
Autoregressive models are constructed based on the product-rule decomposition of joint probability distributions into a sequence of conditional probability distributions,
\begin{equation}
    P(\boldsymbol{\sigma})
    = P(\sigma_{1})P(\sigma_{2}|\sigma_{1})P(\sigma_{3}|\sigma_{<3})...
    \label{eqn:autoreg}
\end{equation}

\begin{figure}[ht]
    \centering
    \includegraphics[width=0.6\textwidth]{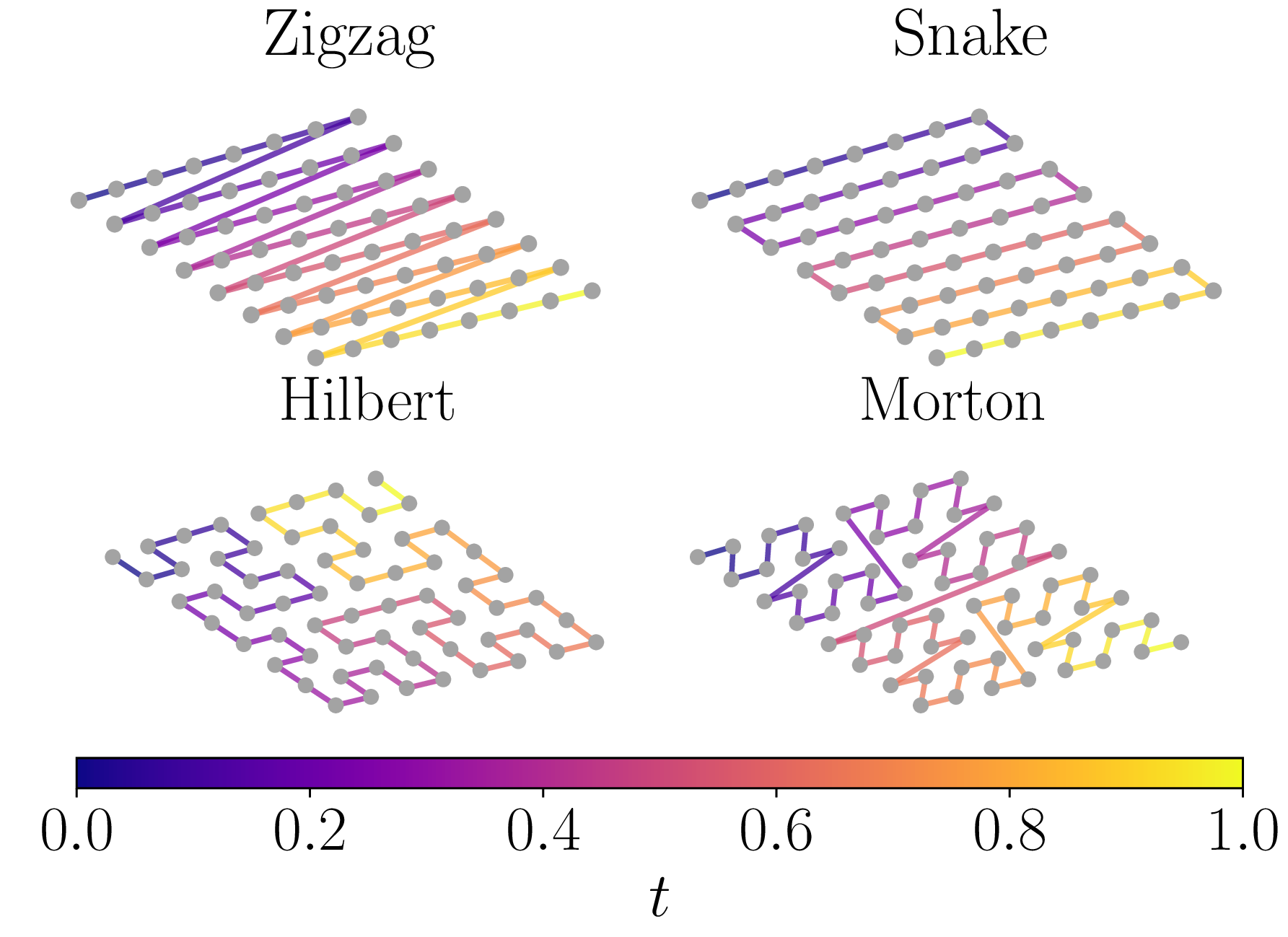}
    \caption{
    Autoregressive paths traversing a 2D square lattice. 
    The color represents the parameter $t$ which is varied, from 0 to 1, to generate the paths from their respective parametric equations.
    }
    \label{fig:autoregpaths}
\end{figure}

This equation holds for any arbitrary ordering of the variables into a sequence.
Therefore, there is a freedom in the assignment of the sequence labels to the sites of a lattice, which generates a {\it path} through the system.
For a 1D lattice, there is a natural method to assign these labels in a jump-free locality-preserving order.
However, for a 2D system, there are multiple ways of constructing paths that exhibit different properties, e.g.~locality-preserving or jump-free properties.
Fig.~\ref{fig:autoregpaths} shows some possible autoregressive paths through a 2D system of $8 \times 8$ sites.
The ``zigzag'', also called a raster scan or sawtooth, is the path is used in many simulation strategies.
Another path that maximizes the length of 1D segments is the ``snake'' path which has also been previously used, e.g.~in Refs.~\cite{moss2023enhancing, Di2022Autoregressive}.
The ``Hilbert'' path is a locality-preserving curve \cite{Hilbert1891ueber,niedermeier_manhattan-distance_1996} that has been demonstrated to have advantages compared to the zigzag path for tensor network methods, such as tree tensor networks (TTN) and matrix product states (MPS) \cite{Cataldi2021hilbert}.
Another famous locality preserving path is the ``Morton'' curve, also known as the Lebesgue curve and the $Z$-order curve \cite{morton1966computer}, traditionally used in resource management in high performance computing \cite{nocentino2010optimizing,lauterbach2009fast, pascucci2001gloabl}.

We focus on two autoregressive architechtures: RNNs and transformers with masked self-attention.
RNNs encode correlations in hidden states which are sequentially passed from one unit to the next.
Thus, long-range correlations in RNNs are encoded via a composition of multiple hidden state passes.
In contrast, transformers encode correlations in the causally masked self-attention mechanism.
In other words, causally allowed uni-directional long-range correlations, consistent with the autoregressive path, are explicitly encoded in the transformer architechture.
transformers and RNNs also differ in their computational cost. transformers scale quadratically with sequence length $N$ and linearly with embedding size $N_\mathrm{e}$, i.e. $\mathcal{O}(N^2 N_\mathrm{e})$  \cite{agrawal2023computational, vaswani2017attention} in contrast to RNNs which have a linear scaling with sequence length and a quadratic scaling in hidden dimension size $N_\mathrm{h}$, i.e $\mathcal{O}(N N_\mathrm{h}^2)$.


\section{Training and results}\label{results}
We train the above autoregressive models using data obtained via thermal Monte Carlo simulations of the Ising model (Eq.~\eqref{Ising}) on a 2D square lattice.
We obtain datasets for multiple linear sizes $L=4,8,16$ and inverse temperatures $\beta$ between 0.286 and 0.667.
Each dataset consist of $10^5$ uncorrelated samples, 80\% of which is used for training and the remainder for validation. Note that these sizes are chosen to be powers of two as the Hilbert and Morton path are traditionally defined on systems of size $2^n$. Note however, there are algorithms to generalize these curves for arbitrary sizes and dimensions \cite{bandoh_address_1999,bandoh_address_2000,kamata_address_1997, hamilton2008compact, rong2021modified, perdacher2020improved}.

The training of the autoregressive models consists of minimizing the Kullback-Leibler divergence (KLD) between the likelihood underlying the dataset and the current state of the model.
Practically, this is equivalent to minimizing the negative log-likelihood (NLL), $\mathcal{L}^{\mathrm{NLL}}$, of obtaining the dataset.
The optimization is performed with ADAM \cite{Kingma2014adam} and a learning rate of $10^{-3}$.

\begin{figure*}[ht]
    \centering
    \includegraphics[width=\textwidth]{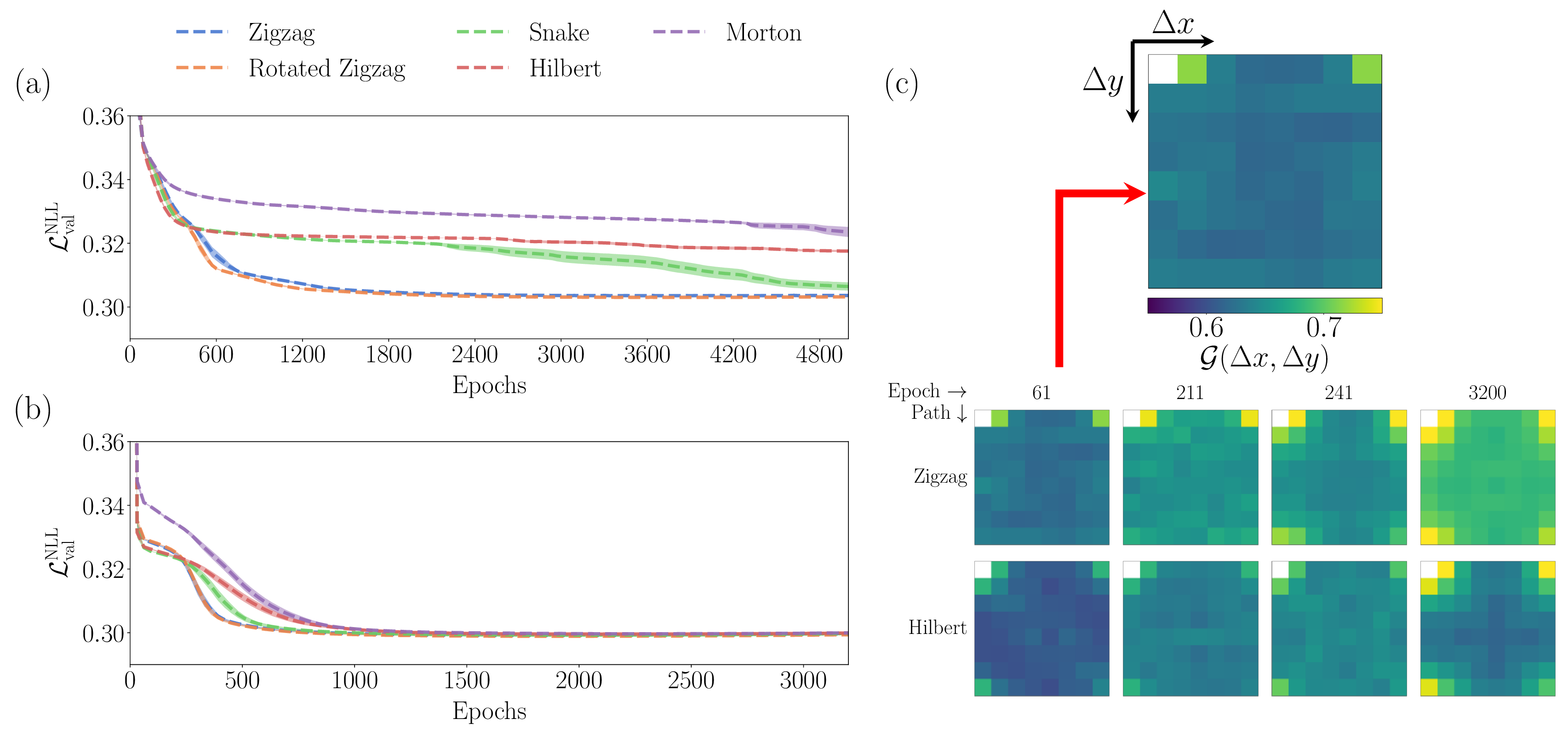}
    \caption{
      Training results for the 2D Ising model near criticality at $\beta = 0.435$
      with linear size $L=8$ for (a) 1D RNN and (b) transformer.
      The critical point of the system is at $\beta_c \approx 0.4407$.
      We train both model architectures with multiple autoregressive paths (listed in Fig.~\ref{fig:autoregpaths}). 
      The shaded area represents the standard error over 10 independent training runs.
      While the transformer is trained with the zigzag and hilbert path, we track the two-point spin-spin correlation function, $\mathcal{G}(\Delta x, \Delta y)$.
      In (c), we plot the two-point spin-spin correlation for when the number of epochs trained is [61, 211, 241, 3200].
    }
    \label{fig:train}
\end{figure*}

To account for the architectural differences between RNNs and transformers, we perform hyperparameters tuning, with the Asynchronous Successive Halving Algorithm (ASHA)\cite{li2020system}, to obtain the optimal parameters for both model architectures.
We found that for the 1D RNN, the optimal hidden state size was $N_h = 16$.
For the transformer, the optimal hyperparameters consist of an embedding size\textbf{} of $N_e = 32$, number of heads $N_{\mathrm{heads}} = 4$, FFNN size $N_{\mathrm{ff}} = 512$, and number of blocks $N_{\mathrm{blocks}} = 2$.

We first study the performance of the models near criticality at $\beta = 0.435$.
The exact critical point of the 2D Ising model is at $\beta_c = \ln \left({1+\sqrt{2}}\right)/2 \approx 0.4407$.
As can be seen in Fig.~\ref{fig:train} (a) and (b), the training performance of each model depends strongly on the autoregressive path formulation.
The paths that perform best for the RNN also perform best for the transformer, however the transformer loss is minimized at a significantly faster rate (in units of epochs).
In particular, the transformer is able to converge relatively rapidly regardless of the autoregressive path, compared to the 1D RNN which requires a large number of epochs in order for the sub-optimal paths to converge.
 
In Fig.~\ref{fig:train} (c), we study the two-point spin-spin correlation function in the models while they are being trained.
The two-point spin-spin correlation function has the following form:
\begin{align}
    \mathcal{G}(\Delta x, \Delta y) = \frac{1}{N} \sum_{x,y} \langle &\sigma_{x,y} \sigma_{x+\Delta x,y+\Delta y} \rangle \nonumber \\
    &- \langle \sigma_{x,y} \rangle \langle \sigma_{x+\Delta x,y+\Delta y}\rangle. \label{eqn:corr}
\end{align}
We observe an anisotropy in the correlations captured by the models with the zigzag and snake path during early-phase training.
In contrast, the correlations in models with the Hilbert and Morton paths are near-isotropic.
This behavior holds for both the 1D RNN and the transformer.
Our results indicate that the slowdown in learning early on for the transformer (before 300 epochs in Fig.~\ref{fig:train} (b)) with the zigzag and snake paths is due to this anisotropy. When the model resolves this anisotropy, it is able to quickly reach the optimal loss.
Finally, it is interesting to observe that even with the anisotropic learning of correlations in the zigzag and snake paths, models with these paths converge faster than the other paths for the case of the Ising model.
This is in contrast to the results with TTN and MPS \cite{Cataldi2021hilbert}, where the Hilbert path was demonstrated to be advantageous.

\begin{figure}[ht]
    \centering
    \includegraphics[width=0.6\textwidth]{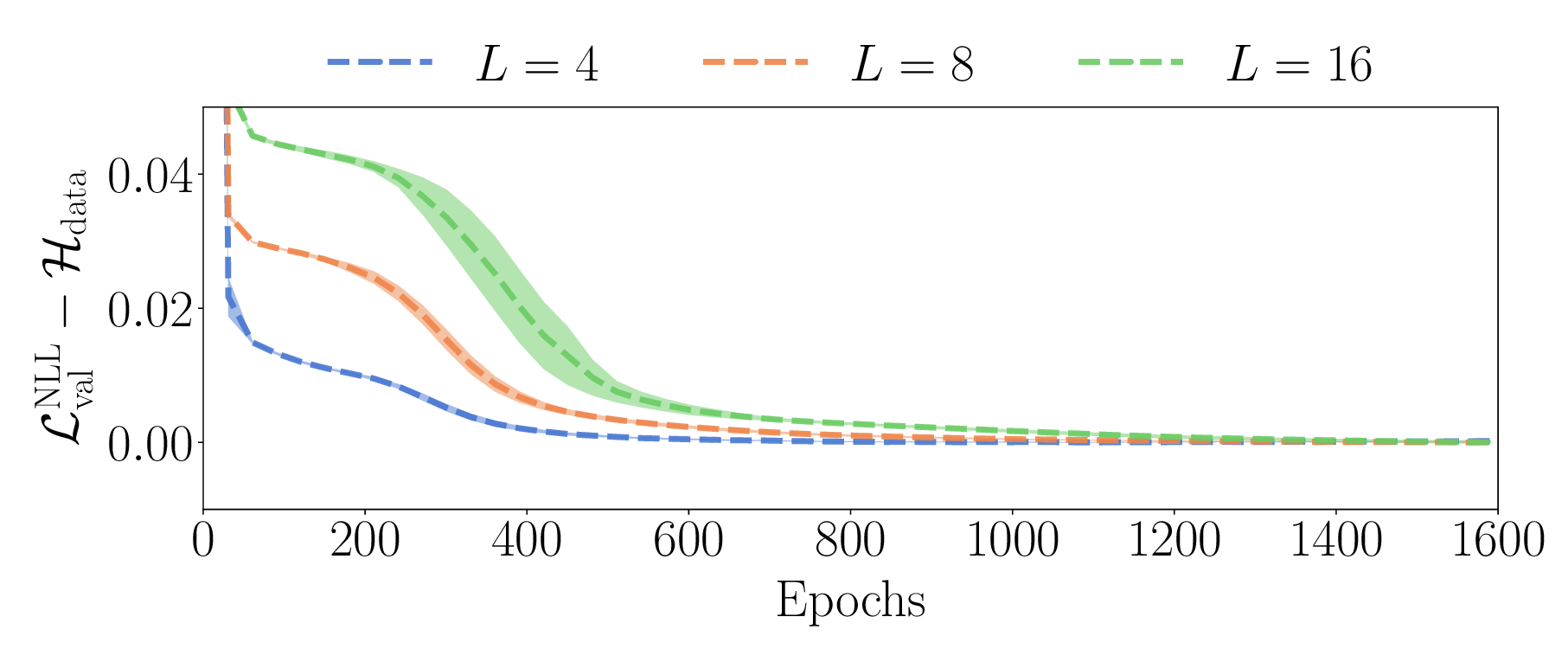}
    \caption{
    Training of the transformer with the zigzag path on the 2D Ising model near criticality ($\beta = 0.435$) with various sizes $L=4,8$ and $16$.
    }
    \label{fig:scale}
\end{figure}

As we scale the transformer, we see that the anisotropic learning of correlations for the zigzag path is consistent across sizes $L=4,8$ and $16$.
We note that the number of epochs required to resolve the anisotropy does not significantly depend on the size of system for the transformer architecture (as shown in Fig.~\ref{fig:scale}).

\begin{figure}[ht]
    \centering
    \includegraphics[width=0.48\textwidth]{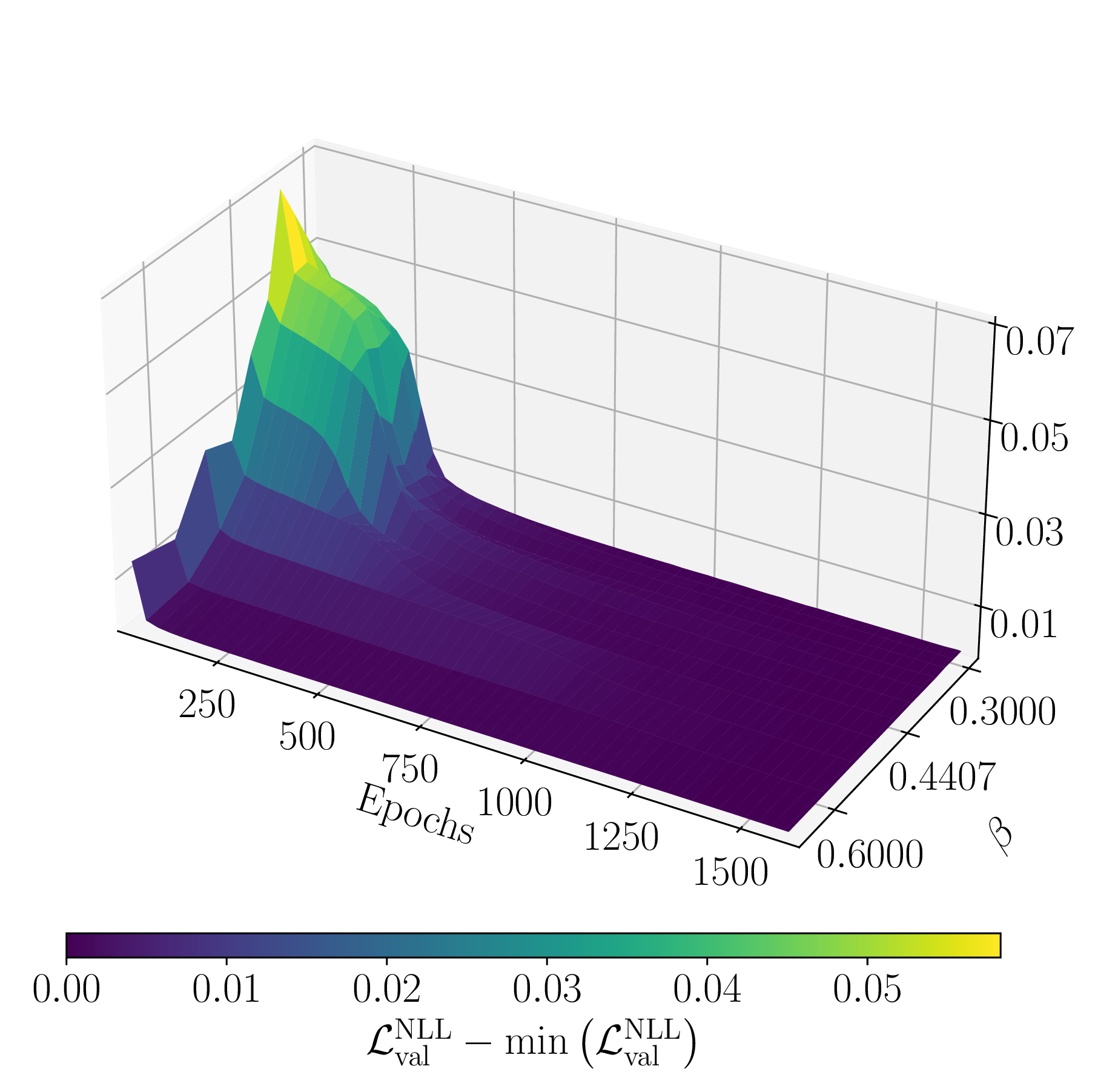}
    \caption{
    Training of the transformer across the 2D Ising model phase transition, $\beta$ between 0.286 and 0.667.
    The theoretical critical point for this system is at $\beta_c \approx 0.4407$.
    }
    \label{fig:transition}
\end{figure}

Finally, we extended our training regime to include a broad range of temperatures across the phase transition of the 2D Ising model.
Due to the algebraic decay of correlations at criticality, training near criticality is the most difficult, supported by Fig.~\ref{fig:transition}.
In Fig.~\ref{fig:transition}, we observe that the initial plateau due to the anisotropic learning of correlations in the models with the zigzag path peaks near the transition and disappears far into the ordered and disordered phases.
This provides evidence for the importance of studying the performance of such models across the phase transitions.

\section{Discussion}\label{discuss}
In this paper, we study the learning behavior of autoregressive generative models when trained on physical data that is not a one-dimensional (1D) sequence.
We focus on two model architectures, recurrent neural networks (RNNs) and transformers.  When trained on data from the two-dimensional Ising model near its critical point, we find a strong dependence of the training  characteristics on the choice of the autoregressive path.

Correlations in the critical Ising model decay algebraically, and therefore represent some of the longest-range interaction of interest in physical systems. 
These correlations are encoded in a sequence (defined by Eq.~\eqref{eqn:autoreg}) in an autoregressive model, which can be any arbitrary 1D path through the lattice.
The important correlations must occur within the {\it context window}, which can span hundreds of thousands of tokens in modern language models.  In RNNs, the context window is defined by the hidden states passed between recurrent units; context is compressed and truncated by multiple nonlinear functions.  In contrast, a transformer encodes long-range context explicitly through the pairwise attention mechanism. 
This means the transformer has direct and complete access to past context and the information can be retrieved via key value search.

Our results show that the selection of the autoregressive path impacts the learning characteristics both recurrent neural networks (RNNs) and transformers.
In the case of a 1D RNN, the locality-preserving paths (Hilbert and Morton) are clearly the worst performers, failing to converge to a good loss value even after significant epochs of training.  This result is surprising, since locality-preserving paths have recently been shown to give superior performance in 1D Matrix Product State (MPS) simulations  \cite{Cataldi2021hilbert}.  
In contrast, for the transformer the final result of the models' training is essentially independent of the autoregressive path.
However, the locality-preserving paths take more epochs to train than the zigzag and snake paths, converging in the same order (zigzag, snake, Hilbert then Morton) as observed in the RNN.
Our results also show that the transformer is able to learn correlations at a significantly faster rate (per epoch) compared to the 1D RNN.
Overall, for the Ising model on a 2D square lattice with periodic boundary conditions, we conclude that the zigzag path performs best out of the various autoregressive paths we tested.
We conjecture that in the comparison between the zigzag and snake path, the zigzag path performs better due to the periodic boundary condition of the model in study.
The periodic boundary contains connections that wrap around the boundary, present in the zigzag path but absent in the snake path.
It would be interesting to repeat our study for models of other geometries, dimensions, and interacting Hamiltonians.

The current work reveals the need to study more closely the effect of autoregressive path choice in the training of modern language models. While our results are for relatively vanilla architectures, in recent years there have been multiple proposals of modified transformers with reduced scaling in sequence length \cite{child2019generating, kitaev2020reformer, katharopoulos2020transformers, lu2021soft, wang2020linformer,krzysztof2021rethinking}.
In certain cases, such models have been demonstrated to outperform the vanilla transformer with significantly fewer resources.
It would be interesting to examine the training efficiency of these models with varying autoregressive paths. Perhaps the combination of modern transformer architechture with a variational principle which optimizes the autoregressive path will lead to increased performance on the expensive training task which faces all large language models.


\bmhead{Acknowledgements}

We would like to acknowledge discussions with Ejaaz Merali and M. Schuyler Moss.
The results in this manuscript were produced with the software packages PyTorch \cite{paszke2019pytorch} and Ray Tune \cite{liaw2018tune}.
We acknowledge support from the Natural Sciences and Engineering Research Council of Canada (NSERC) and the Perimeter Institute for Theoretical Physics. This research was supported in part by grant NSF PHY-2309135 to the Kavli Institute for Theoretical Physics (KITP). Computational support was provided by the facilities of the Shared Hierarchical Academic Research Computing Network (SHARCNET:www.sharcnet.ca) and Compute/Calcul Canada. Research at Perimeter Institute is supported in part by the Government of Canada through the Department of Innovation, Science and Economic Development Canada and by the Province of Ontario through the Ministry of Economic Development, Job Creation and Trade. 

\bmhead{Author contributions}
All analyses and experiments in the manuscript were performed by Y.H.T. under supervision from R.G.M.

\bmhead{Data availability}
The data that support the findings of this study are available from the corresponding author upon reasonable request.



\bibliography{ref}

\end{document}